\documentclass[sigconf]{acmart}

\copyrightyear{2024}
\acmYear{2024}
\setcopyright{acmlicensed}
\acmConference[CIKM '24] {Proceedings of the 33rd ACM International Conference on Information and Knowledge Management}{October 21--25, 2024}{Boise, ID, USA.}
\acmBooktitle{Proceedings of the 33rd ACM International Conference on Information and Knowledge Management (CIKM '24), October 21--25, 2024, Boise, ID, USA}
\acmISBN{979-8-4007-0436-9/24/10}
\acmDOI{10.1145/3627673.3679542}
\usepackage{times}
\usepackage{soul}
\usepackage{url}
\usepackage[utf8]{inputenc}
\usepackage{graphicx}
\usepackage{amsmath}
\usepackage{amsthm}
\usepackage{booktabs}
\usepackage{algorithm}
\usepackage{algorithmic}
\usepackage[switch]{lineno}
\usepackage{xcolor}

\usepackage{amsfonts}
\usepackage{subcaption}
\usepackage{multirow}
\usepackage{array}

\AtBeginDocument{%
  }

\begin{document}

\title[CausalMed]{CausalMed: Causality-Based Personalized Medication Recommendation Centered on Patient Health State}

\author{Xiang Li}
\affiliation{%
  \institution{Yanshan University}
  \city{Qinhuangdao}
  \country{China}
}
\affiliation{
  \institution{Peking University}
  \city{Beijing}
  \country{China}
}
\email{lixiang_222@stumail.ysu.edu.cn}

\author{Shunpan Liang}
\authornote{The corresponding author}
\affiliation{%
  \institution{Yanshan University}
  \city{Qinhuangdao}
  \country{China}
}
\affiliation{%
  \institution{Xinjiang University of Science \& Technology}
  \city{Korla}
  \country{China}
}
\email{liangshunpan@ysu.edu.cn}

\author{Yu Lei}
\affiliation{%
  \institution{Yanshan University}
  \city{Qinhuangdao}
  \country{China}
}
\email{leiyu0160@gmail.com}

\author{Chen Li}
\affiliation{%
  \institution{Yanshan University}
  \city{Qinhuangdao}
  \country{China}
}
\email{lichen36211@gmail.com}

\author{Yulei Hou}
\affiliation{%
  \institution{Yanshan University}
  \city{Qinhuangdao}
  \country{China}
}
\email{ylhou@ysu.edu.cn}

\author{Dashun Zheng}
\affiliation{%
  \institution{Macao Polytechnic University}
  \city{Macao}
  \country{China}
}
\email{p2212871@mpu.edu.mo}

\author{Tengfei Ma}
\authornotemark[1]
\affiliation{%
  \institution{Hunan University}
  \city{Changsha}
  \country{China}
}
\email{tfma@hnu.edu.cn}

\renewcommand{\shortauthors}{Xiang Li, et al.}

\begin{abstract}
Medication recommendation systems are developed to recommend suitable medications tailored to specific patient.
Previous researches primarily focus on learning medication representations, which have yielded notable advances. However, these methods are limited to capturing personalized patient representations due to the following primary limitations: (i) unable to capture the differences in the impact of diseases/procedures on patients across various patient health states; (ii) fail to model the direct causal relationships between medications and specific health state of patients, resulting in an inability to determine which specific disease each medication is treating.
To address these limitations, we propose CausalMed, a patient health state-centric model capable of enhancing the personalization of patient representations.
Specifically, CausalMed first captures the causal relationship between diseases/procedures and medications through causal discovery and evaluates their causal effects.
Building upon this, CausalMed focuses on analyzing the health state of patients, capturing the dynamic differences of diseases/procedures in different health states of patients, and transforming diseases/procedures into medications on direct causal relationships. Ultimately, CausalMed integrates information from longitudinal visits to recommend medication combinations.
Extensive experiments on real-world datasets show that our method learns more personalized patient representation and outperforms state-of-the-art models in accuracy and safety.
\end{abstract}

\ccsdesc[500]{Information systems~Information retrieval}
\ccsdesc[500]{Computing methodologies~Causal reasoning and diagnostics}
\ccsdesc[300]{Applied computing~Health informatics}

\keywords{Medication Recommendation, Causal Inference, Electronic Health Record}

\maketitle

\section{Introduction}
Given the noticeable imbalance in healthcare resources' supply and demand environment, there has been a growing emphasis on AI-based medical systems \cite{background0,background1,background2,background3}. 
Medication recommendation, as one of the domains, aims to provide personalized medication treatment plans for specific patients using data mining \cite{analyze1,analyze2,analyze3} and deep learning \cite{deep1,deep2,deep3} methods.

\begin{figure}
    \centering
    \begin{subfigure}{\linewidth}
        \centering
        \includegraphics[width=\textwidth]{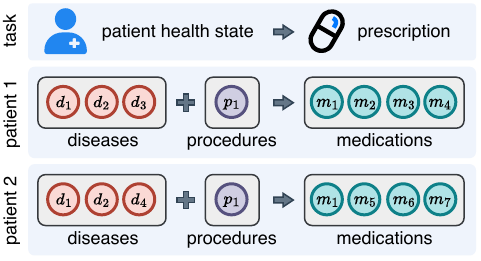}
        \caption{An example: patient 1 has disease \(d_3\), and Patient 2 has disease \(d_4\), their health states differ only in one disease, yet the prescribed medications differ significantly.}
        \label{fig:intro1}
    \end{subfigure}
    \begin{subfigure}{\linewidth}
        \centering
        \includegraphics[width=\textwidth]{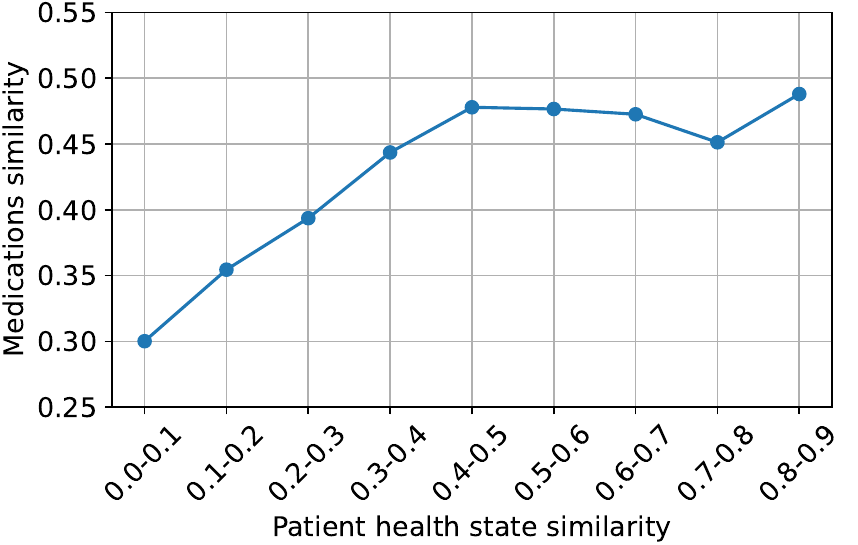}
        \caption{The relationship between the similarity of the patient's health state and the similarity of the medications taken by the patients.}
        \label{fig:intro2}
    \end{subfigure}
    \caption{The problem of insufficient personalization of medication recommendation.}
    \Description{}
    \label{fig:intro}
    \vspace{-0.5cm}
\end{figure}

The term "health state" refers to the sum of all diseases and procedures within a single clinical visit. 
Figure \ref{fig:intro} illustrates two real-life cases, explaining why personalized medication recommendation is crucial.
In our study, we observed a common phenomenon, as depicted in Figure \ref{fig:intro1}, where even if two patients may have highly similar health states, the required medications could be vastly different.
Furthermore, we conducted pairwise comparisons of all medical records in MIMIC-III \cite{mimic3}, as shown in Figure \ref{fig:intro2}, where the results indicate that among patients with medical record similarity ranging from 80\% to 90\%, the average similarity of prescribed medications was only 48.8\%.
In conclusion, due to factors such as individual patient differences, doctors often propose different treatment plans for similar diseases/procedures. Hence, personalized representations tailored to each patient are particularly important.

Previous research methods, such as \cite{micron,cognet}, emphasize the importance of prior medications, suggesting that current medication combinations be recommended based on medications prescribed during the last visit. Meanwhile, other methods, like \cite{safedrug,molerec}, advocate starting from a molecular perspective to construct more accurate representations of medications. However, these studies are primarily medication-centric and limited to addressing the aforementioned issue of personalization due to the following limitations:

(1) \textbf{Failing to model a direct causal relationship between medications and patient's specific health state}:
Most medications are specifically designed for one or two particular diseases/procedures, which means that the relationships between diseases/procedures and medications should be point-to-point causal relationships. Traditional methods typically match disease/procedure sets with medication sets, and construct set-to-set relationships based on co-occurrence. However, these co-occurrence-based methods generate excessive false correlations and fail to capture the direct causal relationship between medications and the specific patient health state, making it impossible to determine which specific disease each medication is treating.

(2) \textbf{Overlooking the difference of diseases in various health states}:
The same disease may assume different roles in various patient health states: it may be the primary cause of other diseases during one visit and a secondary disease resulting from other diseases during another visit. This intricate pathological relationship among diseases implies that the impact of diseases on patients varies across different health states. A similar issue exists with procedures. In dealing with multiple diseases/procedures, existing methods simply sum them with equal weight, failing to consider the varying roles that diseases play in different health states.

To address the aforementioned limitations, based on causal inference \cite{causal_inference1}, we propose a patient health state-centric model named CausalMed. 
Specifically, first, \textbf{to address limitation (1)}, we replace co-occurrence relationships with causal relationships. Through causal discovery, we eliminate backdoor paths \cite{backdoor3,backdoor2} between medical entities (diseases, procedures, medications). Subsequently, we construct corresponding graph networks by estimating the quantified therapeutic effects of specific medications on specific diseases. Ultimately, we transform the initially chaotic set-to-set relationships into clear point-to-point relationships, modeling the direct causal relationships between medications and specific patient states.
At the same time, \textbf{to address limitation (2)}, we introduce a novel approach named Dynamic Self-Adaptive Attention (DSA). Grounded in the causal-based pathological relationships of each health state, DSA learns the variability in the effects that medical entities have on patients under different health states, leading to the construction of personalized patient representations.
Ultimately, by applying this method to historical medical visit data, we integrate longitudinal visit information, resulting in tailored medication recommendations. 

Detailed case study can be found in subsection \ref{subsection:Casestudy}.
Our source code is publicly available on GitHub{\footnote{{\url{https://github.com/lixiang-222/CausalMed}}}}.
We list our main contributions as follows:
\begin{itemize}
    \item \textbf{Important Discovery}:
    We make the first discovery that different diseases express significantly different meanings under various patient health states, and current works fail to capture which specific disease each medication is treating. These factors contribute to the low level of personalization in existing medication recommendations.

    \item \textbf{Novel Framework}:
    We propose a novel medication recommendation framework centered on patients' clinical health state. Utilizing causal inference, we capture point-to-point relationships between diseases and medications and develop a Dynamic Self-Adaptive Attention (DSA) mechanism to discern the dynamic differences in diseases across various health states, thereby enhancing the personalization of patient representations.

    \item \textbf{Extensive Evaluation}:
    Through extensive experiments on real-world datasets, we demonstrate our model's ability to capture personalized patient representations, outperforming other state-of-the-art baselines significantly.

    
    
    
\end{itemize}

\section{Related Work}

\subsection{Medication Recommendation}
Early works in medication recommendation, such as LEAP \cite{leap}, are instance-based, focusing on single-visit data and treating medication recommendation as a multi-instance, multi-label classification task.

The first category of research models sequential relationships in patients' medical records to enhance personalization. RETAIN \cite{retain} personalizes disease predictions using patient histories. MICRON \cite{micron} optimizes therapy by adjusting medications for new symptoms. COGNet \cite{cognet} translates diseases/procedures to medications and reuses effective past treatments.

The second category improves prescription accuracy by capturing relationships between entities. DMNC \cite{dmnc} uses dynamic memory and deep learning to recommend precise medications. Studies like \cite{drug_package,drug_package_generation,megacare} use graph networks to describe connections between medical entities. StratMed \cite{stratmed} stratifies relationships to alleviate data sparsity.

The third category addresses safety and reliability in medication combinations by incorporating partial medication-related knowledge. SafeDrug \cite{safedrug} and MoleRec \cite{molerec} use molecular data to reduce adverse DDIs. Carmen \cite{carmen} integrates patient history with molecular learning to differentiate similar molecules by function.

Our proposed model employs a patient health state-centric framework and enhances personalized patient representations through the incorporation of causal inference methods.

\begin{figure*}
    \centering
    \includegraphics[width=\textwidth]{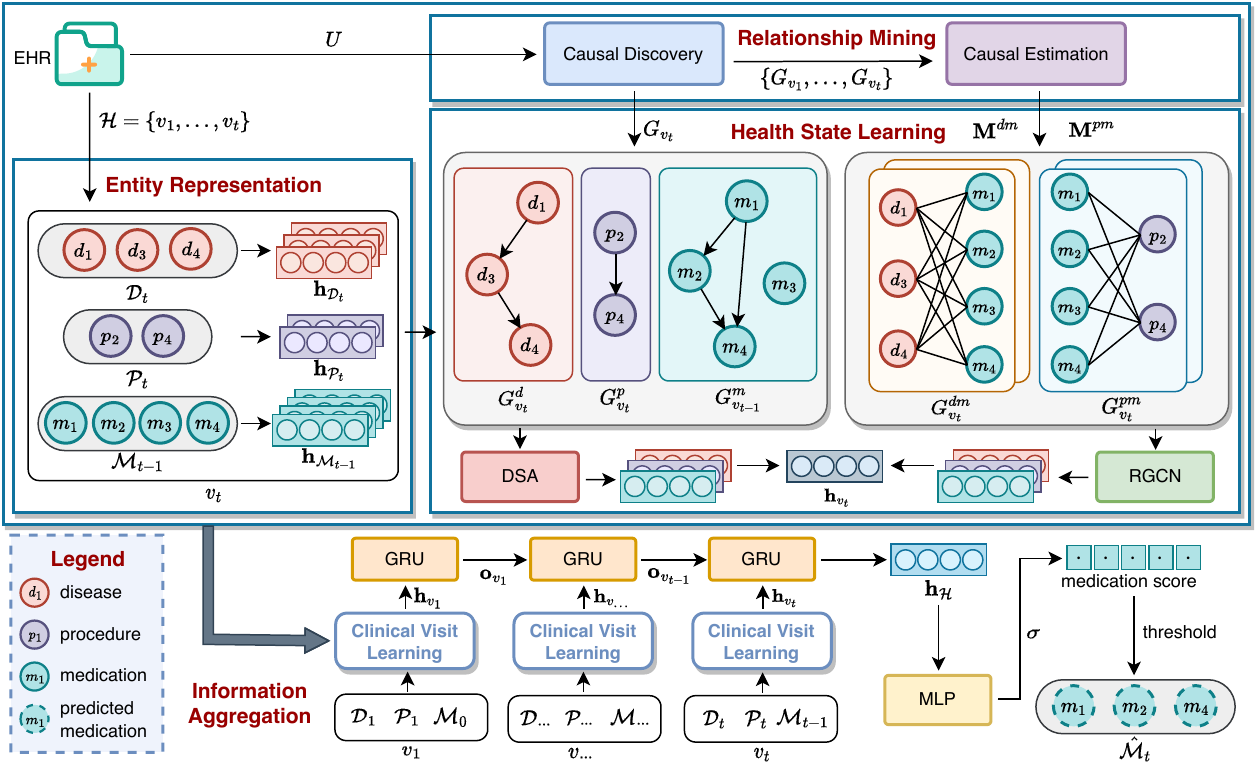}
    \caption{CausalMed framework: The upper section illustrates the process of learning the patient's representation from a single clinical visit. The lower section represents the integration of information from multiple visits and the prediction of medication combinations.}
    \label{fig:model_graph}
    \Description{}
\end{figure*}

\subsection{Causal Inference in Recommender Systems}
The incorporation of causal inference \cite{causal_inference2} into recommender systems (RS) is a novel concept, with ongoing research falling into three main categories.

The first category focuses on tackling data bias. Methods like the backdoor criterion \cite{bias1,bias2,bias3} aim to eliminate confounding factors, addressing challenges such as popularity and exposure bias.

The second category, exemplified by \cite{missing1,missing2,missing3}, couples real-world data collection with inverse probability weighting and counterfactual methods to fill data gaps or reduce noise by introducing counterfactual interactions.

The third category aims to enhance model interpretability and fairness. Studies like \cite{interpret1,interpret2} introduce causal discovery and counterfactual approaches in RS to address the opacity issues inherent in deep learning models.

Our method uses causal inference to establish clear, quantifiable causal relationships, effectively overcoming the limitations of traditional methods that rely on co-occurrence relationships.

\section{Problem Definition}

\subsection{Medical Entity}
A medical entity refers to specific medical concepts that constitute medical data. We use \(\mathcal{D} = \{d_1, d_2, \ldots\}\), \(\mathcal{P} = \{p_1, p_2, \ldots\}\), and \(\mathcal{M} = \{m_1, m_2, \ldots\}\) to represent three sets of medical entities: diseases, procedures, and medications.

\subsection{DDI Matrix}
Our Drug-Drug Interaction (DDI) data is extracted from the Adverse Event Reporting Systems \cite{AERS}. We represent DDI information using a binary matrix \({\mathbf{M}^{ddi}} \in \{0,1\}^{|\mathcal{M}| \times |\mathcal{M}|} \), where \(\mathbf{M}^{ddi}_{ij} = 1\) indicates the presence of an interaction between medication \(m_i\) and medication \(m_j\). A high frequency of DDI suggests potential safety issues in recommended results.

\subsection{Causal Discovery and Estimation}
Causal discovery and estimation constitute two pivotal aspects of causal inference \cite{causal_inference1}. Causal discovery entails employing statistical tools and machine learning algorithms to unveil causal relationships between variables. Causal estimation underscores the quantification of causal effects between variables, relying on observed data.

\subsection{Input and Output}
The model ingests Electronic Health Records (EHR) of patients as its primary input. Each patient's record, symbolized as \(\mathcal{H}\), involves multiple visits \(\mathcal{H} = \{v_1, v_2, \ldots, v_t\}\). Each visit record \(v_t\) encapsulates a tripartite dataset: diseases, procedures, and medications, expressed as \(v_t = \{\mathcal{D}_t, \mathcal{P}_t, \mathcal{M}_t\}\), where \(\mathcal{D}_t\), \(\mathcal{P}_t\), and \(\mathcal{M}_t\) are all multi-hot encoded using 0 and 1.
The output of model, denoted as \(\hat{\mathcal{M}}_t\), is a predicted medication combination for \(v_t\).

\section{The Proposed Model: CausalMed}

As depicted in Figure \ref{fig:model_graph}, our model starts by extracting entity representations from a single clinical visit. In the relationship mining stage, we employ causal discovery and estimation to uncover pathological relationships among medical entities and quantify the point-to-point relationships. 
During the phase of health state learning, we capture the dynamic changes of entities in different patient states through causal relationships, updating and aggregating embeddings of all medical entities to generate visit representations.
Finally, in the information integration stage, we predict medication combinations based on longitudinal patient visit records.

\subsection{Entity Representation}
In this stage, we encode entities within a single clinical visit. Specifically, we start by extracting the current disease \(\mathcal{D}_t\), current procedure \(\mathcal{P}_t\), and the previous medication \(\mathcal{M}_{t-1}\). Following that, we construct embedding tables for diseases, procedures, and medications, represented as \(\mathbf{E}_d \in \mathbb{R}^{|\mathcal{D}| \times dim}\), \(\mathbf{E}_p \in \mathbb{R}^{|\mathcal{P}| \times dim}\), and \(\mathbf{E}_m \in \mathbb{R}^{|\mathcal{M}| \times dim}\) respectively, where \(dim\) represents the embedding dimension and each row in the table retains a vector for specific medical entity. Through the tables, we obtain embeddings for all entities.
\begin{equation}
    \mathbf{h}_{d_i} = \mathbf{E}_d(d_i), \quad \mathbf{h}_{p_j} = \mathbf{E}_p(p_j), \quad \mathbf{h}_{m_k} = \mathbf{E}_m(m_k),
\end{equation} 
where \(d_i \subset \mathcal{D}_t\), \(p_j \subset \mathcal{P}_t\), and \(m_k \subset \mathcal{M}_{t-1}\) represent specific medical entities, \(\mathbf{h}_{d_i} \in \mathbb{R}^{dim}\) represents the embedding for entity \(d_i\), and \(\mathbf{h}_{p_j} \in \mathbb{R}^{dim}\) and \(\mathbf{h}_{m_k} \in \mathbb{R}^{dim}\) follow the same logic for entities \(p_j\) and \(m_k\), respectively.

\subsection{Relationship Mining}

\begin{figure}
    \centering
    \includegraphics[width=\linewidth]{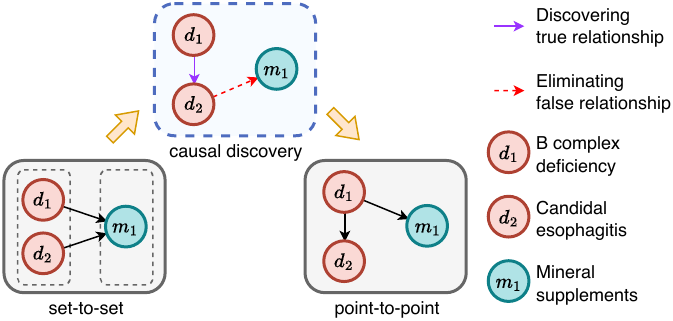}
    \caption{An example of causal discovery, discovering the true relationship between $d_1$ and $d_2$ and eliminating the false relationship between $d_2$ and $m_1$, ultimately transforming set-to-set relationships into point-to-point relationships.}
    \label{fig:causal}
    \Description{}
    \vspace{-0.5cm}
\end{figure}

At this stage, we use causal discovery and estimation to uncover genuine relationships among medical entities, eliminating spurious co-occurrence relationships. We transform set-to-set relationships into precise point-to-point relationships and refine causal effects for each group.

Firstly, as illustrated in Figure \ref{fig:causal}, the co-occurrence-based method simply associates all diseases and medications that appear together in a single medical visit, assuming that $d_1$ and $d_2$ are both related to $m_1$. However, causal discovery methods can learn that $d_2$ is caused by $d_1$ by resolving backdoor paths \cite{backdoor1,backdoor2}, and there is no actual relationship between $m_1$ and $d_2$.
We utilize causal discovery to learn the causal structure from set-to-set relationships, uncovering the true underlying relationships and eliminating spurious ones caused by co-occurrence, ultimately generating point-to-point relationships between diseases and medications. 

Specifically, we analyze the EHR to obtain the distribution \(U\) of medical entities and employ a variation of Greedy Intervention Equivalence Search (GIES) \cite{GIES,cdt} to identify confounding bias and generate a causal relationship graph \(G\). GIES optimizes the Bayesian equivalence class. Concerning the Bayesian equivalence class, we define the scoring criterion as \(S(G, U)\) to evaluate the quality of each causal graph \(G\) learned from the distribution \(U\). 
\begin{gather}
    S(G, U) = \sum_{i=1}^{n} s(X_i, P^G_{a_i}),\\
    G' = \text{GIES}(S,G),
\end{gather}
where \(n\) is the number of variables, \(X_i\) is the variable including disease, surgery, and medication, and \(P^G_{a_i}\) represents the parents of \(X_i\) in the graph \(G\). 
And $s(\cdot, \cdot)$ refers to the method of obtaining the equivalence class from \(X_i\) and its parent nodes (\(P^G_{a_i}\)), through which we can produce the initial graph \(G\) and continuously update it.
The $\text{GIES}(\cdot)$ optimizes and learns an optimized graph \(G'\) from the initial graph \(G\) and the equivalence score \(S\). We apply the above process to each medical visit, generating a series of pathological graphs \(\mathcal{G} = \{G_{v_1}, G_{v_2}, \ldots\}\) that encompass all visits.

Subsequently, adopting the principles of causal estimation, we define diseases and procedures as treatment variables and medications as outcome variables. We use a generalized linear model adjusted based on the backdoor criterion \cite{backdoor1,backdoor3} to calculate the quantified effects of medications on diseases and procedures. Ultimately, we obtain a quantified relationship between each disease/procedure and each medication, and construct matrices of causal effects, denoted as \(\mathbf{M}^{dm} \in \mathbb{R}^{|\mathcal{D}| \times |\mathcal{M}|}\) and \(\mathbf{M}^{pm} \in \mathbb{R}^{|\mathcal{P}| \times |\mathcal{M}|}\).

\subsection{Health State Learning}
We use causal relationships to learn patients' health states. Homomorphic relationships update entity meanings under different health states, while heterogeneous relationships apply disease/procedure-medication causality. This information is integrated to create a personalized clinical visit representation.

\subsubsection{Homomorphic Relationship Learning}
As shown in the left half of the Health State Learning part of Figure \ref{fig:model_graph}, we learn entity representations through homomorphic relationships. By exploring pathological relationships based on causality, we capture the dynamic differences between entities, enhancing our ability to capture personalized patient characteristics.

In different health states, a disease's impact on patients can vary. In one state, a disease may cause other diseases, while in another, it may result from them. We introduce a Dynamic Self-Adaptive Attention (DSA) mechanism to learn the role of diseases in various health states. Using causal positioning to determine a disease's role in the current clinical setting, we assign different weights, enhancing personalized patient representations.

\begin{figure}
    \centering
    \includegraphics[width=\linewidth]{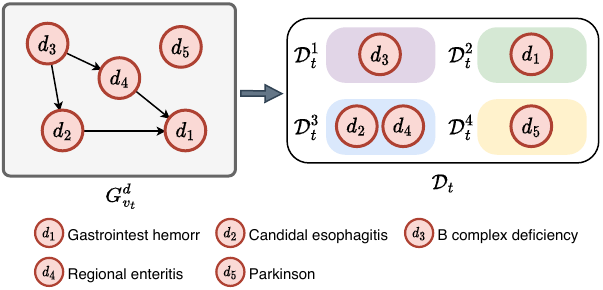}
    \caption{Using the classifier within DSA, we analyze the causal position of each node in the causal graph \( G^d_{v_t} \) to gauge its influence on the patient's current state. For instance, if \( d_3 \) is identified as a causal disease, it is classified as \( {D}^1_t \).}
    \label{fig:dsa}
    \Description{}
    \vspace{-0.7cm}
\end{figure}

First, we extract relationships from the causal graph \(G_{v_t}\) involving entities from the current visit and construct three isomorphic causal graphs: \(G^d_{v_t}\), \(G^p_{v_t}\), and \(G^m_{v_{t-1}}\). We then categorize the patient's diseases/procedures as shown in Figure \ref{fig:dsa}. For diseases \(\mathcal{D}_t\), we categorize them into four types \(\mathcal{D}^j_t\) based on \(G^d_{v_t}\): 1) Causal diseases \(\mathcal{D}^1_t\), which initiate other diseases and represent the primary ailment; 2) Effect diseases \(\mathcal{D}^2_t\), influenced by other diseases and indicative of secondary symptoms; 3) Middle diseases \(\mathcal{D}^3_t\), both influencing and influenced by other diseases; 4) Independent diseases \(\mathcal{D}^4_t\), existing without direct causal links to other diseases.

Since each patient has a unique disease history, the same disease may occupy different positions in the causal graphs \(G^d_{v_t}\) for different patients, leading to different group classifications. The DSA mechanism learns the dynamic changes of disease groups across various health states, capturing personalized differences. It updates entity representations based on these groupings to enhance the expressiveness of key entities.
The specific formula is as follows:
\begin{gather}
    \mathcal{D}^j_t = \text{Classify}(d_i, G^d_{v_t}),\\
    \mathbf{h}^r_{d_i} = \mathbf{h}_{d_i} \cdot w^j_t,\\
    w^j_t = \frac{\exp(\mathbf{W} \cdot \mathbf{h}^{D^j}_t + b)}{\sum_{k=1}^{4} \exp(\mathbf{W} \cdot \mathbf{h}^{D^k}_t + b)},
\end{gather}
Where \(\text{Classify}(\cdot)\) assigns \(d_i\) to a specific category \(\mathcal{D}^j_t\), the DSA generates dynamic weight \(w^j_t\). \(\mathbf{W}\) and \(b\) are trainable weight matrix and bias term, \(\mathbf{h}^{D^j}_t\) represents the sum of embeddings of diseases within \(\mathcal{D}^j_t\), and \(\mathbf{h}^r_d\) is the modified embedding of diseases. The same approach is extended to procedures and medications, yielding \(\mathbf{h}^r_p\) and \(\mathbf{h}^r_m\).

\subsubsection{Heterogeneous Relationship Learning}
As shown in the right half of the Health State Learning part of Figure \ref{fig:model_graph}.
At this stage, we update the representations of medical entities based on the correlations between diseases/procedures and medications obtained through causal estimation in relationship mining. This process aims to enhance the accuracy of entity representations.

We utilize the causal effect matrices \(\mathbf{M}^{dm}\) and \(\mathbf{M}^{pm}\) to construct disease-medication and procedure-medication complete bipartite graphs \(G^{dm}_{v_t}\) and \(G^{pm}_{v_t}\) for single visit. 

Taking  \(G^{dm}_{v_t}\)  as an example, the nodes encompass both \(\mathcal{D}_t\) and \(\mathcal{M}_{t-1}\).  
By utilizing Relational Graph Convolutional Networks (RGCN) \cite{RGCN}, we categorize edges corresponding to relationships with similar causal effects into the same type. Subsequently, we establish point-to-point relationships between diseases and medications. The representation of \(d_i\) at layer \(l\) is then updated on the point-to-point relationship using the following formula:
 \begin{gather}
     \mathbf{h}_{d_i}^{l+1} = \sigma (\mathbf{W}_0^l \cdot \mathbf{h}_{d_i}^l + \sum_{r \in \mathcal{R}} \mathbf{W}_r^l \cdot (\frac{1}{c_{{d_i},r}} \sum_{j \in \mathcal{N}_r({d_i})} \mathbf{h}_j^l)),\\
     \mathbf{W}_r^l = I + \Theta_r^l,
\end{gather} 
Where \(\mathbf{W}_r^l\) is the weight matrix for relation \(r\) at layer \(l\), \(\mathbf{W}_0^l\) is the weight matrix for the node itself, \(\mathcal{N}_r({d_i})\) is the set of neighboring nodes of \({d_i}\) for relation \(r\), and \(c_{{d_i},r}\) is a normalization factor. \(I\) is the identity matrix, and \(\Theta_r^l\) is the weight update matrix for relation \(r\) at layer \(l\). In \(G^{pm}_{v_t}\), we repeat this process to obtain entity embeddings \(\mathbf{h}^e_d\), \(\mathbf{h}^e_p\), and \(\mathbf{h}^e_m\).

\subsubsection{Aggregation of Entity Information}
We use the residual method to stack the embeddings after initialization and weighting. By aggregating the updated embeddings with element-wise addition, we obtain three set representations: \(\mathbf{h}_{\mathcal{D}_t}\), \(\mathbf{h}_{\mathcal{P}_t}\), and \(\mathbf{h}_{\mathcal{M}_{t-1}}\). These representations are then concatenated with the representation \(\mathbf{h}_{v_t}\) for the current clinical visit.
\begin{gather}
    \begin{gathered}
    \mathbf{h}_{\mathcal{D}_t}=\text{Agg}(\mathbf{h}^r_d,\mathbf{h}^e_d),
    \mathbf{h}_{\mathcal{P}_t}=\text{Agg}(\mathbf{h}^r_p,\mathbf{h}^e_p),\\
    \mathbf{h}_{\mathcal{M}_{t-1}}=\text{Agg}(\mathbf{h}^r_m,\mathbf{h}^e_m),
    \end{gathered}\\
    \mathbf{h}_{v_t} = [\mathbf{h}_{\mathcal{D}_t}||\mathbf{h}_{\mathcal{P}_t}||\mathbf{h}_{\mathcal{M}_{t-1}}].
\end{gather}    

\subsection{Medication Recommendation}
To deeply mine personalized patient representations, we consider the patient's medical history. Each medical record is represented and integrated to recommend medication combinations. Multiple GRUs capture temporal dependencies among visits, generating GRU output $\mathbf{o}_{v_t}$ and patient embedding $\mathbf{h}_\mathcal{H}$. An MLP layer with activation function $\sigma$ converts this into medication scores. Medications $\hat{m}_i$ with scores exceeding $\delta$ are chosen.
\begin{gather}
    \mathbf{h}_\mathcal{H} = \mathbf{o}_{v_t} = \text{GRU}(\mathbf{o}_{v_{t-1}},\mathbf{h}_{v_t}),\\
    score = \sigma (\text{MLP}(\mathbf{h}_\mathcal{H})),\\
    \hat{m}_i = 
    \begin{cases}
        1 ,&  score_{m_i} \ge \delta  \\
        0 ,&  score_{m_i} < \delta 
    \end{cases}.
\end{gather}    

\begin{table*}[h]
    \centering
    \caption{The performance of each model on the test set regarding accuracy and safety. The best and the runner-up results are highlighted in bold and underlined respectively under t-tests, at the level of 95\% confidence level.}
    \begin{tabular}{|*{1}{>{\centering\arraybackslash}p{1.4cm}}| *{4}{>{\centering\arraybackslash}p{1.1cm}} *{1}{>{\centering\arraybackslash}p{1.25cm}} | *{4}{>{\centering\arraybackslash}p{1.1cm}} *{1}{>{\centering\arraybackslash}p{1.25cm}} |}
    \toprule
    \multirow{2}{*}{Model}
    & \multicolumn{5}{c|}{MIMIC-III} & \multicolumn{5}{c|}{MIMIC-IV} \\ 
    \cmidrule(lr){2-6} \cmidrule(lr){7-11}
    & Jaccard$\uparrow$ & DDI$\downarrow$ & F1$\uparrow$ & PRAUC$\uparrow$ & Avg.\#Med & Jaccard$\uparrow$ & DDI$\downarrow$ & F1$\uparrow$ & PRAUC$\uparrow$ & Avg.\#Med \\
    \midrule
    LR          & 0.4924    & 0.0830    & 0.6490    & 0.7548    & 16.0489   & 0.4569    & 0.0783    & 0.6064    & 0.6613    & 8.5746 \\
    ECC         & 0.4856    & 0.0817    & 0.6438    & 0.7590    & 16.2578   & 0.4327    & 0.0764    & 0.6129    & 0.6530    & 8.7934 \\
    RETAIN      & 0.4871    & 0.0879    & 0.6473    & 0.7600    & 19.4222   & 0.4234    & 0.0936    & 0.5785    & 0.6801    & 10.9576 \\ 
    LEAP        & 0.4526    & 0.0762    & 0.6147    & 0.6555    & 18.6240   & 0.4254    & 0.0688    & 0.5794    & 0.6059    & 11.3606 \\
    GAMENet     & 0.4994    & 0.0890    & 0.6560    & 0.7656    & 27.7703   & 0.4565    & 0.0898	& 0.6103	& 0.6829    & 18.5895 \\
    SafeDrug    & 0.5154    & \textbf{0.0655}       & 0.6722    & 0.7627    & 19.4111   & 0.4487	& \textbf{0.0604}	& 0.6014	& 0.6948 & 13.6943 \\
    MICRON      & 0.5219    & 0.0727    & 0.6761    & 0.7489    & 19.2505    & 0.4640    & 0.0691    & 0.6167    & 0.6919    & 12.7701  \\
    COGNet      & \underline{0.5312}    & 0.0839    & 0.6744    & 0.7708    & 27.6335    & \underline{0.4775}	& 0.0911	& 0.6233	& 0.6524 & 18.7235 \\
    MoleRec     & 0.5293    & 0.0726    & \underline{0.6834}    & \underline{0.7746}    & 22.0125      & 0.4744	& 0.0722	& \underline{0.6262}	& \underline{0.7124} & 13.4806  \\
    \midrule
    \textbf{CausalMed}   & \textbf{0.5389}   & \underline{0.0709}  & \textbf{0.6916}   & \textbf{0.7826}    & 20.5419         & \textbf{0.4899}	 & \underline{0.0677}	 & \textbf{0.6412}	   & \textbf{0.7338} & 14.4295\\
    \bottomrule
    \end{tabular}

    \label{tab:comparison}
    \vspace{-0.3cm}
\end{table*}

\subsection{Model Training and Inference}
Our training and inference processes share the same pipeline. In training, we optimize all learnable parameters and define the following loss functions: binary cross-entropy loss $\mathcal{L}_{bce}$, multi-label margin loss $\mathcal{L}_{multi}$, and DDI loss $\mathcal{L}_{ddi}$.
\begin{gather}
    \mathcal{L}_{bce} = -\sum_{i=1}^{|\mathcal{M}|}{m_i}\log({\hat{m}_i})+(1-{m_i})\log (1-{\hat{m}_i}),\\
    \mathcal{L}_{multi} = \sum_{i,j:{m_i}=1,{m_j}=0} \frac{\max(0,1-({\hat{m}_i}-{\hat{m}_j}))}{|\mathcal{M}|},\\
    \mathcal{L}_{ddi} = \sum_{i=1}^{|\mathcal{M}|}\sum_{j=1}^{|\mathcal{M}|} \mathbf{M}^{ddi}_{ij}\cdot{\hat{m}_i}\cdot {\hat{m}_j}.
\end{gather}    

We adopt a loss aggregation approach that aligns with the methodology used in prior work \cite{molerec}.
\begin{gather}
    \mathcal{L} = \alpha (\beta \mathcal{L}_{bce}+(1-\beta)\mathcal{L}_{multi})+(1-\alpha )\mathcal{L}_{ddi},\\
    \alpha =     
    \begin{cases}
        1 ,& \text{DDI rate}\le \gamma   \\
        \max\{0, 1- \frac{\text{DDI rate}-\gamma}{kp}\} ,& \text{DDI rate}> \gamma
    \end{cases},
\end{gather}  
where $\beta$ is hyperparameters, and the controllable factor $\alpha$ is relative to DDI rate, $\gamma \in (0,1)$ is a DDI acceptance rate and $kp$ is a correcting factor for the proportional signal.


\section{Experiments}
In this section, we conduct extensive experiments aimed at answering the following key research questions (RQ):
\begin{itemize}
    \item \textbf{RQ1}: Does CausalMed provide more accurate and safe recommendations than state-of-the-art medication recommendation systems?
    \item \textbf{RQ2}: Can the two core modules proposed in this paper enhance the effectiveness of recommendations?
    \item \textbf{RQ3}: Are the causal relationships used by CausalMed more appropriate for medication recommendations than the co-occurrence relationships used in previous medication recommendations?
    \item \textbf{RQ4}: Why CausalMed can generate more personalized patient representations based on causality?
    \item \textbf{RQ5}: Can the representation process of CausalMed be made transparent, and are its results interpretable?
\end{itemize}

\subsection{Experiments Setup}

\begin{table}
    \centering
    \caption{Statistics of the datasets.}
    \begin{tabular}{|c|c|c|}
    \toprule
    Item & MIMIC-III & MIMIC-IV \\
    \midrule
    \# patients         & 6,350  & 60,125 \\
    \# clinical events  & 15,032 & 156,810 \\
    \# diseases         & 1,958  & 2,000 \\
    \# procedures       & 1,430  & 1,500 \\
    \# medications      & 131   & 131\\
    avg. \# of visits    & 2.37  & 2.61\\
    avg. \# of medications & 11.44  & 6.66\\
    \bottomrule
    \end{tabular}
    \label{tab:datasets}
    \vspace{-0.3cm}
\end{table}

\subsubsection{Datasets}
We verify the model performance on the MIMIC-III \cite{mimic3} and MIMIC-IV \cite{mimic4}. 
Our methodology for data processing and evaluation mirrored that of \cite{safedrug}, encompassing: 1) Data preprocessing, with outcomes detailed in Table \ref{tab:datasets}; 2) Division of datasets into training, validation, and test sets in a 4/6-1/6-1/6 ratio; 3) A sampling approach involving the extraction of test data from the test set via ten rounds of bootstrap sampling.

\subsubsection{Evaluation Metrics}
1)\textbf{Jaccard}: measures the overlap between predicted and actual sets, indicating similarity;
2) \textbf{DDI-rate}: Assesses safety by indicating the potential for harmful medication interactions. A lower value signifies a safer medication combination;
3)\textbf{F1-Score}: Balances precision and recall, a generic evaluation metric for recommendation tasks;
4)Precision-Recall AUC (\textbf{PRAUC}): Represents the model's ability to distinguish between classes across various thresholds.
5)\textbf{Avg.\#Med}: Reflects the average number of medications recommended per clinical visit and is used for reference only, not indicative of the recommendation system's overall quality.

\begin{table*}
    \centering
    \caption{The performance of multiple variants of CausalMed on evaluation metrics. The best and the runner-up results are highlighted in bold and underlined respectively under t-tests, at the level of 95\% confidence level.}
    \begin{tabular}{|*{1}{>{\centering\arraybackslash}p{3cm}}| *{4}{>{\centering\arraybackslash}p{1.3cm}} | *{4}{>{\centering\arraybackslash}p{1.3cm}}|}
    \toprule
    \multirow{2}{*}{Model}
    & \multicolumn{4}{c|}{MIMIC-III} & \multicolumn{4}{c|}{MIMIC-IV} \\ 
    \cmidrule(lr){2-5} \cmidrule(lr){6-9}
    & Jaccard$\uparrow$ & DDI$\downarrow$ & F1$\uparrow$ & PRAUC$\uparrow$ & Jaccard$\uparrow$ & DDI$\downarrow$ & F1$\uparrow$ & PRAUC$\uparrow$ \\
    \midrule
    CausalMed w/o T     & \underline{0.5369}    & 0.0734  & \underline{0.6900}    & \underline{0.7824}    & \underline{0.4878} & 0.0702 & \underline{0.6405} & \underline{0.7330} \\
    CausalMed w/o P	    & 0.5324    & \underline{0.0731}    &  0.6860     & 0.7745  & 0.4838 & \underline{0.0695} & 0.6355 & 0.7258 \\
    CausalMed w/o T+P   & 0.5339    & 0.0740                & 0.6873   & 0.7788   & 0.4847 & 0.0711 & 0.6373 & 0.7309 \\ 
    \midrule
    \textbf{CausalMed}   & \textbf{0.5389}   & \textbf{0.0709} & \textbf{0.6916}   & \textbf{0.7826}  & \textbf{0.4899}	 & \textbf{0.0677} & \textbf{0.6412}	   & \textbf{0.7338}\\
    \bottomrule
    \end{tabular}

    \label{tab:ablation}
\end{table*}

\subsubsection{Baselines}
\textbf{LR} \cite{lr} is a classifier for medication recommendation but doesn't consider historical impacts on predictions.
\textbf{ECC} \cite{ecc} improves multi-label classification by splitting it into multiple binary problems.
\textbf{RETAIN} \cite{retain} uses a reverse time attention mechanism to highlight influential historical events in sequential medical data.
\textbf{LEAP} \cite{leap} uses reinforcement learning to optimize medication strategies by predicting label dependencies and DDI.
\textbf{GAMENet} \cite{gamenet} employs a graph-enhanced memory network to enhance the safety and efficacy of medication combinations.
\textbf{SafeDrug} \cite{safedrug} predicts DDI by analyzing structural similarities between medication molecules.
\textbf{MICRON} \cite{micron} uses a residual network to personalize medication recommendations by updating the latest combinations.
\textbf{COGNet} \cite{cognet} generates medication sets using cloning and prediction mechanisms, reintroducing significant past medications.
\textbf{MoleRec} \cite{molerec} analyzes molecular substructures to better predict medication properties and interactions.

\subsubsection{Implementation Details}
In our study, as shown in subsection \ref{subsection:Parameters}, we configure embedding dimensions at \( \text{dim} = 64 \). The RGCN architecture consists of a 2-layer graph neural network segmented into 5 groups. For GRU and MLP, a single-layer network with a 64-dimensional hidden layer and ReLU activation is employed. The parameters, consistent with previous work \cite{molerec}, include a dropout rate of 0.5, a threshold \( \delta = 0.5 \), loss function parameters \( \beta = 0.95 \), \( kp = 0.05 \), and an acceptance rate \( \gamma = 0.06 \). Training uses the Adam Optimizer with a learning rate of \( lr = 0.0005 \) and \( L2 \) regularization with a coefficient of 0.005. Experiments are conducted on an Ubuntu machine with 30GB memory, 12 CPUs, and a 24GB NVIDIA RTX3090 GPU.

\subsection{Performance Comparison (RQ1)}

To demonstrate the effectiveness of CausalMed, we conducted comparative experiments with several state-of-the-art baselines, and the results are shown in Table \ref{tab:comparison}. In the realm of traditional machine learning-based methods, both LR and ECC demonstrate sub-optimal performance in terms of accuracy and safety.
Despite employing deep learning strategies, LEAP fails to consider patients' longitudinal historical visit information, resulting in no significant improvement in outcomes. Although RETAIN and GAMENet effectively improve accuracy, they ignore the consideration of DDI issues. SafeDrug introduces the importance of molecules and uses molecular characterization to identify potential DDI problems, greatly improving safety. However, it fails to balance accuracy and safety, performing poorly in terms of accuracy. MICRON proposes using the residual network to learn the difference in patient status between two consecutive visits. It does not directly prescribe medication but updates the last prescription. The results are greatly improved. COGNet treats this task as a translation task and uses Transformer to further improve efficiency. However, due to insufficient consideration of safety, there are still certain flaws. MoleRec conducts further research on molecules, using the substructure of molecules to learn medication representations and recommendations, improving in multiple dimensions.

CausalMed introduces a health state-centric framework, recommending personalized medication combinations based on causality. Experimental results demonstrate a significant enhancement of our proposed method compared to other baselines, validating the effectiveness of CausalMed.

\begin{table*}
    \centering
    \caption{The impact of three hyperparameters on the model. The best and the runner-up results are highlighted in bold and underlined respectively under t-tests, at the level of 95\% confidence level. Parameters used in the final model are in bold.}
    \begin{tabular}{|*{1}{>{\centering\arraybackslash}p{3cm}}| *{4}{>{\centering\arraybackslash}p{1.3cm}} | *{4}{>{\centering\arraybackslash}p{1.3cm}}|}
    \toprule
    parameter & \multicolumn{4}{c|}{MIMIC-III} & \multicolumn{4}{c|}{MIMIC-IV} \\ 
    \midrule
    \textbf{Edge type}   & 4 & \textbf{5} & 6 & 7 & 4 & \textbf{5} & 6 & 7 \\
    \cmidrule(lr){1-1} \cmidrule(lr){2-5} \cmidrule(lr){6-9}
    Jaccard$\uparrow$  & \underline{0.5377}    & \textbf{0.5389}    & 0.5354    & 0.5363 & \underline{0.4856} & \textbf{0.4899} & 0.4834 & 0.4846\\
    DDI$\downarrow$    & 0.0710    & \underline{0.0709}    & \textbf{0.0706}    & 0.0713 & 0.0695 & \underline{0.0677} & \textbf{0.0670} & 0.0708\\
    \midrule
    \textbf{RGCN layer}        & 1         & \textbf{2}         & 3         & 4  & 1         & \textbf{2}         & 3         & 4\\
    \cmidrule(lr){1-1} \cmidrule(lr){2-5} \cmidrule(lr){6-9}
    Jaccard$\uparrow$  & 0.5356    & \textbf{0.5389}    & 0.5349    & \underline{0.5368} & \underline{0.4874} & \textbf{0.4899} & 0.4837 & 0.4854\\
    DDI$\downarrow$    & \textbf{0.0707}    & \underline{0.0709}    & 0.0711    & 0.0724 & \underline{0.0698} & \textbf{0.0677} & 0.0691 & 0.0701\\
    \midrule
    \textbf{Embedding dim}     & 32        & \textbf{64}        & 128       & 256 & 32        & \textbf{64}        & 128       & 256\\
    \cmidrule(lr){1-1} \cmidrule(lr){2-5} \cmidrule(lr){6-9}
    Jaccard$\uparrow$  & \underline{0.5371}    & \textbf{0.5389}    & 0.5349    & 0.5311 & \underline{0.4862} & \textbf{0.4899} & 0.4823 & 0.4806\\
    DDI$\downarrow$    & 0.0723    & \textbf{0.0709}    & \underline{0.0716}    & 0.0732 & \textbf{0.0676} & \underline{0.0677} & 0.0707 & 0.0723\\
    \bottomrule
    \end{tabular}
    \label{tab:parameter}
\end{table*}

\vspace{-0.2cm}
\subsection{Ablation Study (RQ2\&RQ3\&RQ4)}



To verify the effectiveness of our proposed innovations, we thoroughly evaluate three variant models. CausalMed $w/o$ T omits the causal effects and the refinement of the point-to-point relationships between diseases/procedures and medications. CausalMed $w/o$ P excludes pathological relationships, thereby not identifying disease differences. CausalMed $w/o$ T+P represents the model with the concurrent removal of both modules. Since the ablation versions originate from the same framework, we remove the Avg.\#Med which only served as a reference.

As illustrated in Table \ref{tab:ablation}, the result indicates that compared to CausalMed $w/o$ T+P, both CausalMed $w/o$ T and CausalMed $w/o$ P show significant improvements in accuracy. 
This indicates that CausalMed $w/o$ T can identify causal point-to-point associations between diseases/surgeries and medications, providing a more accurate relational basis through the graph network process. At the same time, CausalMed $w/o$ P can learn personalized representations based on the health state of the patient.
Notably, CausalMed $w/o$ P exhibits a significantly higher DDI rate. This is attributed to our modeling of inter-medication relationships during the health state learning stage, where we learn about the causal interactions between medications, thereby enhancing the safety of the outcomes.
The higher performance of CausalMed indicates that the two modules have a synergistic effect when used together, leading to a more pronounced enhancement. 
Through the construction of these variant models, we can effectively verify the contributions of the different innovation points to our overall model.

\vspace{-0.3cm}
\subsection{Parameters Analysis}
\label{subsection:Parameters}
In this subsection, we investigate the impact of several hyperparameters. These parameters include the number of types of edges in the bipartite graph, the number of layers in the RGCN, and the embedding dimensions for each medical entity. We select four neighboring configurations and utilize the Jaccard and DDI rate to represent the accuracy and safety of the results in Table \ref{tab:parameter}.



Our analysis indicates that optimal and balanced outcomes are achieved with a configuration of 5 edge types, 2 RGCN layers, and 64 dimensions. The number of edge types delineates the granularity of relationship categorization, where a limited number can result in too broad divisions, reducing the uniqueness of embeddings, and excess may lead to smaller sample sizes per category, compromising the embeddings' representativeness. The two-layer RGCN facilitates the integration of the two node types within the graph, while additional layers could obscure the demarcation among embeddings. In terms of dimension, low dimensions risk insufficient training, and excessively high dimensions could impinge on both accuracy and safety, owing to overfitting.

\vspace{-0.3cm}
\subsection{Quality Study (RQ3)}
\label{subsection:QualityStudy}

\begin{figure}
    \centering
    \begin{subfigure}{\linewidth}
        \raggedright
        \includegraphics[width=\textwidth]{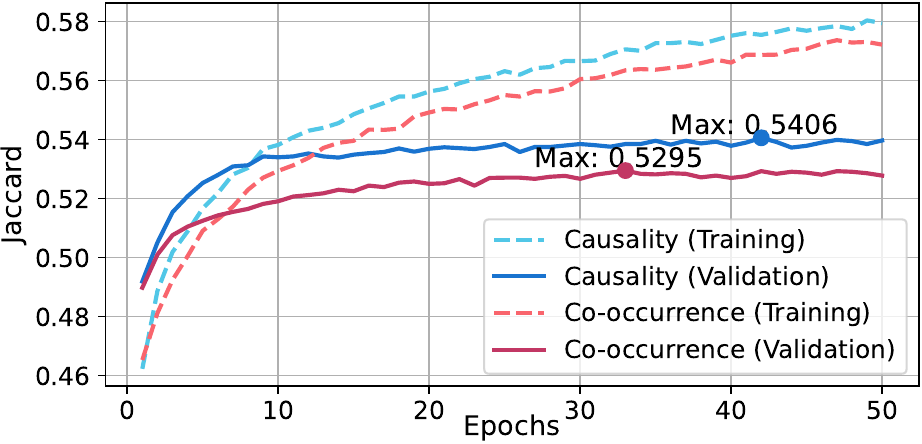}
        \caption{Results of the co-occurrence relationship-based model on the training and validation sets.}
        \label{fig:quality1}
    \end{subfigure}
    \begin{subfigure}{\linewidth}
        \raggedright
        \includegraphics[width=\textwidth]{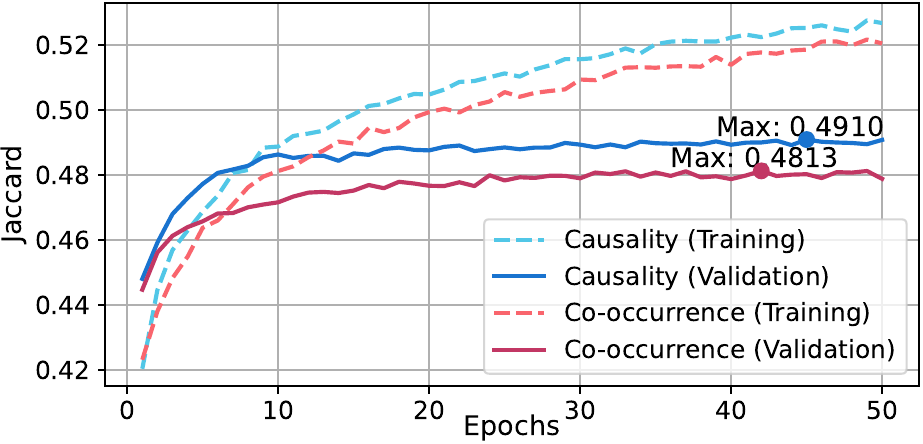}
        \caption{Results of causality-based models on training and validation sets.}
        \label{fig:quality2}
    \end{subfigure}
    \caption{The performance of co-occurrence-based and causality-based methods on Jaccard.}
    \label{fig:quality}
    \Description{}
    \vspace{-0.5cm}
\end{figure}

To validate the significance of introducing causality, we compare the performance of causal relationships and co-occurrence relationships in our model.
We create a comparative model by replacing all causal relationships in CausalMed with co-occurrence relationships. 
Specifically, we remove causal inference methods and replace the causal effects between diseases/procedures and medications with co-occurrence rates. 
Since co-occurrence methods cannot capture the pathological relationships between medical entities, we remove an entire process of health state learning and fuse them using equal-weight addition. 
The subsequent recommendation methods remain the same as the original model. 

The results for both methods on training and validation sets are illustrated in Figure \ref{fig:quality}, indicating that the co-occurrence-based method is inferior to the causality-based method, whether on the training or validation set. 
Firstly, the causal effects between diseases and medications modeled through causal estimation are characterized by point-to-point, offering finer details than co-occurrence rates derived from set-to-set relationships. 
Furthermore, methods based on causal relationships can dynamically construct patient representation, whereas methods based on co-occurrence simply aggregate multiple entity embeddings. 
Therefore, methods based on causality are more suitable for medication recommendation compared to those based on co-occurrence relationships.

\subsection{Case Study (RQ3\&RQ4\&RQ5)}
\label{subsection:Casestudy}

\begin{figure}
    \centering
    \begin{subfigure}{\linewidth}
        \centering
        \includegraphics[width=0.8\textwidth]{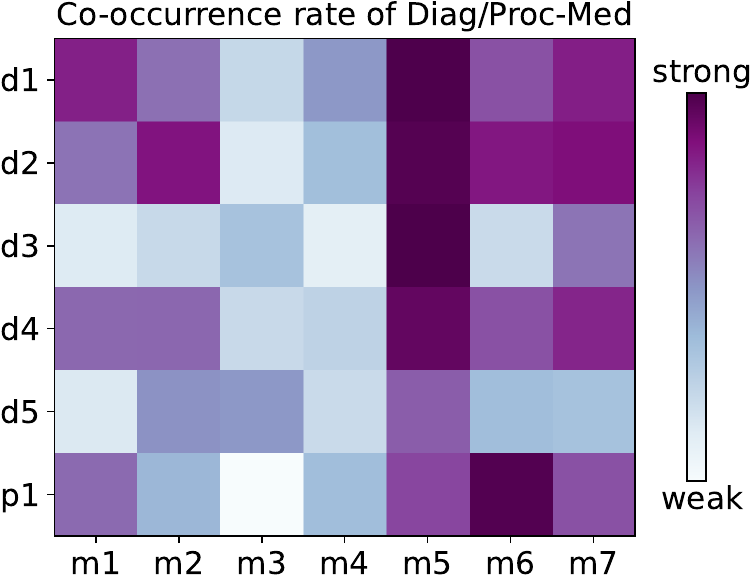}
        \caption{Quantitative relationships between specific diseases/adverse reactions and medications based on co-occurrence relationships.}
        \label{fig:case_study1.1}
    \end{subfigure}
    
    \begin{subfigure}{\linewidth}
        \centering
        \includegraphics[width=0.8\textwidth]{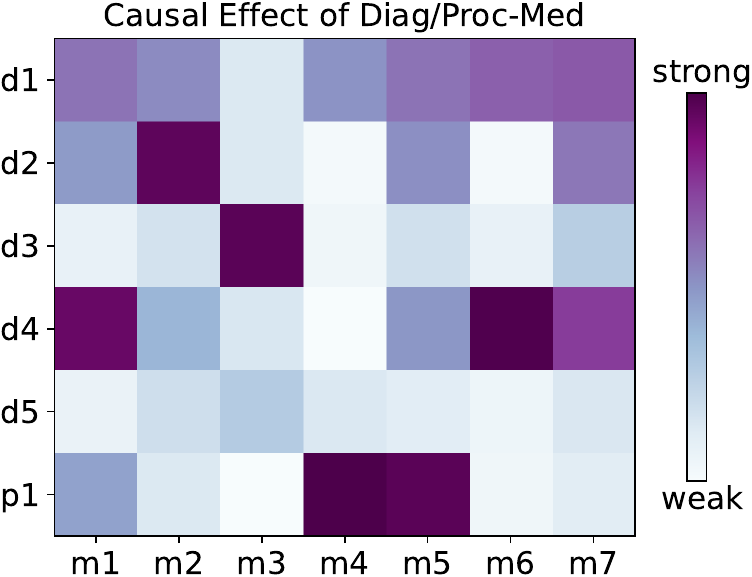}
        \caption{Quantitative relationships between specific diseases/adverse reactions and medications based on causal relationships.}
        \label{fig:case_study1.2}
    \end{subfigure}
    
    \begin{subfigure}{\linewidth}
        \raggedright
        \includegraphics[width=\textwidth]{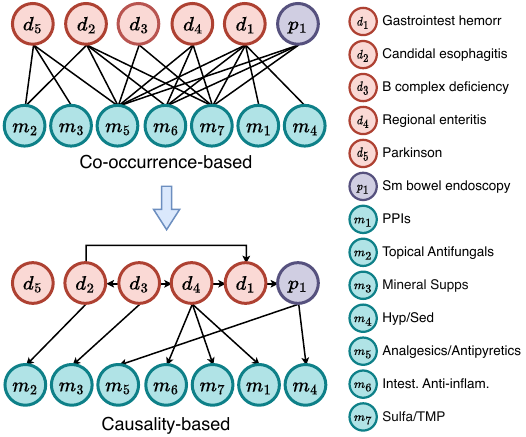}
        \caption{Prescription principles for a specific visit.}
        \label{fig:case_study2}
    \end{subfigure}
    
    \caption{The learning process under co-occurrence-based and causality-based approaches from the real clinical sample.}
    \label{fig:case_study}
    \Description{}
\end{figure}

To more intuitively demonstrate how causality-based methods capture personalized patient representations, we employ the same co-occurrence-based comparative model as in the previous subsection and make transparent the learning process of patient representations in a real case under both approaches. This ultimately proves the rationality and interpretability of the recommended results in this study.
Figure \ref{fig:case_study} is derived from a clinical visit associated with gastric bleeding. Sub-figure \ref{fig:case_study1.1} and \ref{fig:case_study1.2} illustrate the quantified relationships between diseases/procedures and medications, while Sub-figure \ref{fig:case_study2} provides a visualization of the learning results.

Sub-figure \ref{fig:case_study1.1} shows co-occurrence-based relationships between diseases/procedures and medications, which are ambiguous. Medications like $m_5$ (analgesics), $m_6$ (intestinal anti-inflammatory agents), and $m_7$ (sulfonamides) are frequently prescribed and show high co-occurrence rates with many diseases. In contrast, rare medications like $m_3$ (mineral supplements) have low co-occurrence rates, making specific associations difficult to determine. As depicted in the upper half of Sub-figure \ref{fig:case_study2}, this results in common medications being associated with nearly every disease/procedure while rare medications fail to establish appropriate relationships. Consequently, the model preferentially recommends common medications, overlooking rare ones, sometimes leading to positive feedback during training and amplifying the error.

In contrast, relationships based on causality are clearer, as shown in Sub-figure \ref{fig:case_study1.2}. Point-to-point causal estimation links each medication to its specific diseases or procedures, avoiding spurious correlations found in co-occurrence-based relationships. The lower half of Sub-figure \ref{fig:case_study2} shows that by analyzing differences in patients' diseases/procedures, we create dynamic patient profiles. By integrating precise disease/procedure-medication relationships, we build more personalized patient representations. Our method learns that doctors only need to prescribe for \(d_4\) (Regional enteritis) and \(p_1\) (Small bowel endoscopy) related to \(d_1\) (Gastrointest hemorr), without prescribing specifically for \(d_1\). In contrast, the co-occurrence-based method prescribes all these medications for \(d_1\).

We demonstrate how causality-based methods enhance medication recommendation personalization and validate our proposed framework's reasonability, transparency and interpretability via comparative analysis.

\section{Conclusion}
We introduce a causality-based, patient health state-centric framework named CausalMed, which effectively addresses the issues of ambiguous relationships between diseases/procedures and medications and the challenge of capturing differences among diseases/ procedures, thereby significantly enhancing the personalization of patient representations. Extensive experiments on real-world datasets validate the efficacy of the proposed framework.

\begin{acks}
This work was supported by the Innovation Capability Improvement Plan Project of Hebei Province (No. 22567637H), the S\&T Program of Hebei(No. 236Z0302G), and HeBei Natural Science Foundation under Grant (No.G2021203010 \& No.F2021203038).
\end{acks}

\bibliographystyle{ACM-Reference-Format}
\bibliography{CausalMed}

\end{document}